\documentclass[sigconf]{acmart}

\usepackage{booktabs} 
\usepackage{fancyhdr}
\usepackage{enumitem}
\usepackage{subcaption}
\usepackage{multirow}
\usepackage{listings}
\usepackage{xcolor}
\usepackage{amsmath}

\usepackage{mathrsfs,amssymb}
\usepackage[boxed,ruled]{algorithm2e}
\SetKwRepeat{Do}{do}{while}
\usepackage{makecell}
\usepackage[para,online,flushleft]{threeparttable}
\usepackage{mathtools}
\usepackage{bm} 

\DeclareMathAlphabet{\mathcal}{OMS}{cmsy}{m}{n}
\usepackage{setspace}
\usepackage{booktabs} 
\setcopyright{rightsretained}

\copyrightyear{2019} 
\acmYear{2019} 
\setcopyright{acmcopyright}
\acmConference[FPGA '19]{The 2019 ACM/SIGDA International Symposium on Field-Programmable Gate Arrays}{February 24--26, 2019}{Seaside, CA, USA}
\acmBooktitle{The 2019 ACM/SIGDA International Symposium on Field-Programmable Gate Arrays (FPGA '19), February 24--26, 2019, Seaside, CA, USA}
\acmPrice{15.00}
\acmDOI{10.1145/3289602.3293904}
\acmISBN{978-1-4503-6137-8/19/02}

\vspace{-10.5in}

\usepackage{kantlipsum}
\makeatletter
\def\@maketitle{\newpage
 \null
 \setbox\@acmtitlebox\vbox{%
\baselineskip 20pt
\vskip -2cm                  
  \begin{center}
    {\ttlfnt \@title\par}       
    \vskip -2em                
{\subttlfnt \the\subtitletext\par}\vskip 1.25em
    {\baselineskip 18pt\aufnt   
     \lineskip 0em             
     \begin{tabular}[t]{c}\@author
     \end{tabular}\par}
    \vskip 0em               
  \end{center}}
 \dimen0=\ht\@acmtitlebox
 \unvbox\@acmtitlebox
 \ifdim\dimen0<0.0pt\relax\vskip-\dimen0\fi}

\begin{document}
\title{REQ-YOLO: A Resource-Aware, Efficient Quantization Framework for Object Detection on FPGAs}



\author{Caiwen Ding$^{2,+}$,  Shuo Wang$^{1,+}$, Ning Liu$^2$, Kaidi Xu$^2$, Yanzhi Wang$^2$ and Yun Liang$^{1,3,*}$}
\affiliation{$^+$These authors contributed equally.}
\thanks{$^*$Corresponding author.}
\affiliation{$^1$Center for Energy-Efficient Computing \& Applications (CECA), School of EECS, Peking University, China \\ \institution{$^2$Department of Electrical \& Computer Engineering, Northeastern University, Boston, MA, USA \\ $^3$Peng Cheng Laboratory, Shenzhen, China}
{$^1$\{shvowang, ericlyun\}@pku.edu.cn}, $^2$\{ding.ca, liu.ning, xu.kaid\}@husky.neu.edu, $^2$yanz.wang@northeastern.edu}

\renewcommand{\shortauthors}{C. Ding et al.}

\begin{abstract}
Deep neural networks (DNNs), as the basis of object detection, will play a key role in the development of future autonomous systems with full autonomy. The autonomous systems have special requirements of real-time, energy-efficient implementations of DNNs on a power-constrained system.
Two research thrusts are dedicated to performance and energy efficiency enhancement of the inference phase of DNNs. The first one is model compression techniques while the second is efficient hardware implementation. Recent works on extremely-low-bit CNNs such as the binary neural network (BNN) and XNOR-Net replace the traditional floating point operations with binary bit operations which significantly reduces the memory bandwidth and storage requirement. However, it suffers from non-negligible accuracy loss and underutilized digital signal processing (DSP) blocks of FPGAs.

To overcome these limitations, this paper proposes REQ-YOLO, a resource aware, systematic weight quantization framework for object detection, considering both algorithm and hardware resource aspects in object detection. We adopt the block-circulant matrix method and propose a heterogeneous weight quantization using {\em Alternating Direction Method of Multipliers} (ADMM), an effective optimization technique for general, non-convex optimization problems.
To achieve real-time, highly-efficient implementations on FPGA, we present the detailed hardware implementation of block circulant matrices on CONV layers and develop an efficient processing element (PE) structure supporting the heterogeneous weight quantization, CONV dataflow and pipelining techniques, design optimization, and a template-based automatic synthesis framework to optimally exploit hardware resource. 
Experimental results show that our proposed  REQ-YOLO framework can significantly compress the YOLO model while introducing very small accuracy degradation.
\end{abstract}

%
%



\maketitle

\section{Introduction}

Autonomous systems such as unmanned aerial vehicles (UAVs), autonomous underwater vehicles (AUVs), and unmanned ground vehicles (UGVs) have been rapidly growing for performing surveillance, object detection~\cite{bazi2018convolutional}, and object delivery~\cite{high2018apparatus} tasks in scientific, military, agricultural, and commercial applications. The full autonomy of such systems relies on the integration of artificial intelligence software with hardware.

The deployment of deep neural networks (DNNs) in autonomous systems include multiple aspects, i.e., object detection/surveillance algorithms, and advanced control (e.g., deep reinforcement learning technique). Since DNN-based advanced control is not widely in place yet, we focus on the former aspect. Object detection algorithms are different from image classification~\cite{krizhevsky2012imagenet} in that the former need to simultaneously detect and track multiple objects with different sizes. Representative object detection algorithms include R-CNN~\cite{girshick2014rich} and YOLO~\cite{redmon2016you}. The autonomous system applications have special requirements of real-time, energy-efficient implementations on a power-constrained system.

Two research thrusts are dedicated to performance and energy efficiency enhancement of the inference phase of DNNs. The first one is model compression techniques for DNNs~\cite{han2016eie,lin2016fixed,wu2016mobile,courbariaux2015binaryconnect}, including weight pruning, weight quantization, low-rank approximation, etc. S. Han et al.~\cite{han2016eie} have proposed an iterative DNN weight pruning method, which could achieve 9$\times$ weight reduction on the AlexNet model and has been applied to LSTM RNN as well~\cite{han2017ese}. However, this method results in irregularity in weight storage, and thereby degrades the parallelism degree and hardware performance, as observed in~\cite{Wang2018clstm,ding2017circnn,wen2016learning}. 
Recent work~\cite{ding2017circnn,Wang2018clstm} adopts block-circulant matrices for weight representation in DNNs in both image classification DNN~\cite{ding2017circnn}  and LSTM RNN~\cite{Wang2018clstm} tasks. 
This method is demonstrated to achieve higher hardware performance than iterative pruning due to the regularity in weight storage and computation. 
The second one is efficient hardware implementations, including FPGAs and ASICs~\cite{wei2017automated,chen2017eyeriss,zhang2015optimizing,qiu2016going,alwani2016fused,lu2018spwa,lu2017evaluating,wei2018tgpa,xiao2017exploring}. FPGAs are gaining more popularity for striking a balance between high hardware performance and fast development round. A customized hardware solution on FPGA can offer significant improvements in energy efficiency and power consumption compared to CPU and GPU clusters. 

Convolutional (CONV) layers are more computation-intensive than fully-connected (FC) layers. Recently CONV layers are becoming more important in state-of-the-art DNNs~\cite{redmon2017yolo9000,krizhevsky2012imagenet}. 
Extremely-low-bit CNNs such as the binary neural network (BNN)~\cite{courbariaux2015binaryconnect} and XNOR-Net\cite{rastegari2016xnor} have demonstrated hardware friendly ability on FPGAs~\cite{umuroglu2017finn}. Binarization not only reduces memory bandwidth and storage requirement but also replaces the traditional floating point operations with binary bit operations, which can be efficiently implemented on the look-up-tables (LUT)-based FPGA chip, whereas suffering non-negligible accuracy degradation on large datasets due to the over-quantized weight representation. More importantly, the majority of DSP resource will be wasted due to the replacement of multipliers, introducing significant overhead on LUTs. Overall, there lacks a systematic weight quantization framework considering hardware resource aspect on FPGAs. In addition, despite the research efforts devoted to the hardware implementation of image classification tasks~\cite{krizhevsky2012imagenet,lecun2015lenet,he2016deep}, there lacks enough investigation on the hardware acceleration of object detection tasks.

In this paper, we propose REQ-YOLO, a resource-aware, efficient weight quantization framework for object detection by exploring both software and hardware-level optimization opportunities on FPGAs. 
We adopt the block-circulant matrix based compression technique and propose a heterogeneous weight quantization using ADMM on the FFT results considering hardware resource. It is necessary to note that the proposed framework is also applicable to other model compression techniques.
To enable real-time, highly-efficient implementations on FPGA, we present the detailed hardware implementation of block circulant matrices on CONV layers and develop an efficient processing element (PE) structure supporting the heterogeneous weight quantization method, dataflow based pipelining, design optimization, and a template-based automatic synthesis framework. 

Our specific contributions are as follows:
\begin{itemize}

     \item We present a detailed hardware implementation and optimization of block circulant matrices on CONV layers on object detection tasks.

    \item 
    We present a heterogeneous weight quantization method including both equal-distance and mixed powers-of-two methods considering hardware resource on FPGAs. We adopt ADMM to directly quantize the FFT results of weight.

    \item We employ an HLS design methodology for productive development and optimal hardware resource exploration of our FPGA-based YOLO accelerator. 

\end{itemize} 

Experimental results show that our proposed REQ-YOLO framework can significantly compress the YOLO model while introducing very small accuracy degradation. Our framework is very suitable for FPGA and the associated YOLO implementations outperform the state-of-the-art designs on FPGAs.

\section{Preliminaries on Object Detection}

Deep neural networks (DNNs) have dominated the state-of-the-art techniques of object detection. There are typically two main types of object detection methods: (i) region proposal based method and (ii) proposal-free method. For the region proposal based methods, R-CNN first generates potential object regions and then performs classification on the proposed regions~\cite{everingham2010pascal}. SPPnet~\cite{he2014spatial}, Fast R-CNN~\cite{girshick2015fast}, and Faster R-CNN~\cite{ren2015faster} are typical in this category. As for the proposal-free methods, MS-CNN~\cite{cai2016unified} proposes a unified multi-scale CNN for fast object detection. YOLO~\cite{redmon2016you} simultaneously predicts multiple bounding boxes and classification class probabilities. Compared to the region proposal-based methods, YOLO does not require a second classification operation for each region and therefore it achieves significant faster speed. However, YOLO suffers from several drawbacks: (i) YOLO makes a significant number of localization errors compared to Fast R-CNN; (ii) Compared to region proposal-based methods, YOLO has a relatively low recall. To improve the localization and recall while maintaining classification accuracy, YOLO v2~\cite{redmon2017yolo9000} has been proposed. In this paper, we focus on an embedded version of YOLO - tiny YOLO~\cite{darkflow} for hardware implementation. Compared to other versions such as YOLO v2~\cite{redmon2017yolo9000}, v3~\cite{redmon2018yolov3}, and YOLO~\cite{redmon2016you}, it has a smaller network structure and fewer weight parameters, but with tolerable accuracy degradation.

\begin{figure}[t]

\centering

\includegraphics[width = 1\columnwidth]{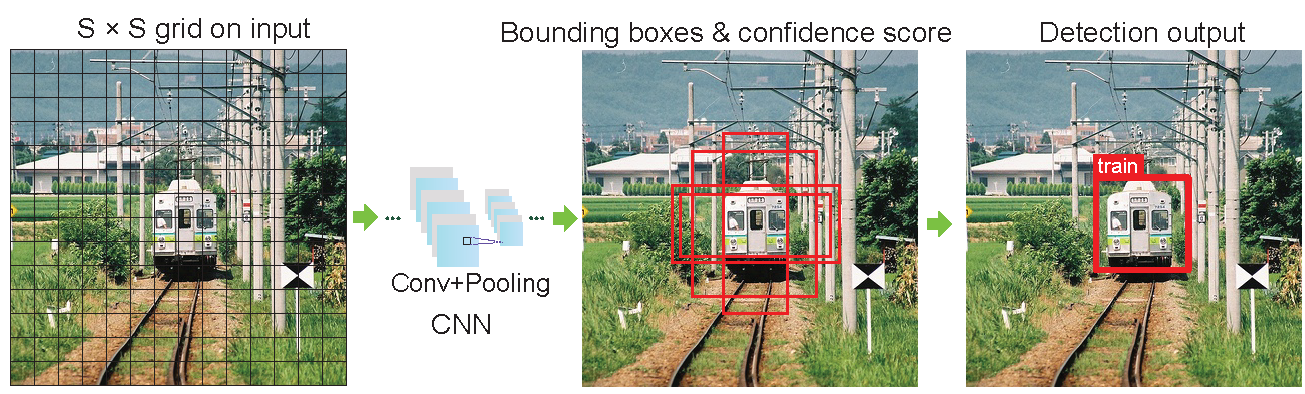}

\caption{ Example of object dection using YOLO.}

\label{fig:grid}

\vspace{-0.2in}

\end{figure}

\begin{figure*}[t]

\centering

\includegraphics[width = 1.9\columnwidth]{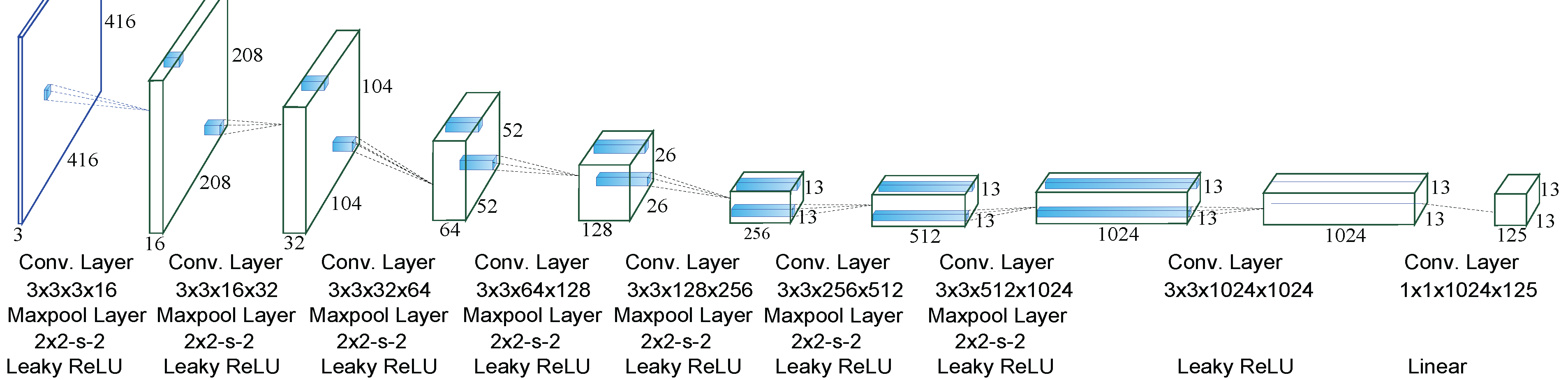}

\caption{The tiny YOLO architecture with convolution layers.}

\label{fig:yolo}

\vspace{-0.2in}

\end{figure*}

\subsection{You Only Look Once (YOLO) Network}

Fig.~\ref{fig:grid} shows an overview of object detection and tiny YOLO application. It uses CONV layers to extract features from images, anchor boxes to predict bounding boxes, and regression for object detection, based on the yolov2 tiny~\cite{darkflow} framework.
The input image (frame) is separated into an $S \times S$ grid. Each grid cell detects an object and predicts $B$ bounding boxes and the corresponding confidence scores when the grid cell and the center of object overlap each other. Typically $S=13$ and $B=5$. For each bounding box, there are 5 predictions made: $x,~ y,~ w,~ h$, and a confidence score. ($x,~y$) is the coordinates of the box center located in the grid cell, and $w$ and $h$ are the width and height of the bounding box. The confidence score is defined as $Pr(Obj) \times IOU_{pred}^{truth}$, where $Pr(Obj)$ is the prediction probability and $IOU_{pred}^{truth}$ is the \emph{intersection over union} ($IOU$). Here, $IOU$ is determined by dividing the area of overlap between the predicted bounding box and its corresponding ground-truth bounding box by the area of union. 

There are $C$ \emph{conditional class probabilities} on the grid cell containing the object, $Pr(Class_i|Obj)$, predicted by each grid cell. The \emph{class-specific confidence scores} for each box are calculated as follows:
\vspace{-0.3cm}
\begin{equation}
   Pr(Class_i|Obj) \cdot Pr(Obj) \cdot IOU_{pred}^{truth}= Pr(Class_i) \cdot IOU_{pred}^{truth} 
\end{equation}

The scores represent how accurate the box is pertinent to the object and the probability of the object class. These predictions are encoded as an $S \times S \times (B \cdot 5+C))$ tensor.
Fig.~\ref{fig:yolo} shows the architecture of tiny YOLO. It has 9 CONV layers. The input images are re-sized to 416 by 416 from the PASCAL 2007 detection dataset~\cite{everingham2010pascal}. From the $1^{st}$ to the $8^{th}$ layer, a $3 \times 3$ CONV operation (stride 1 and zero padding with 1) is followed by a max pooling operation with $2 \times 2$ filters and stride 2. In the last CONV layer, the $1 \times 1$ CONV operation reduces the feature space from the previous layer.

\subsection{Convolutional (CONV) Layers}

Convolutional (CONV) layers convolve each input feature map with an $r \times r$ weight filter. The convolutional results are then accumulated, added with a bias. After passing the intermediate result through a activation function (such as rectified linear unit (ReLU), sigmoid, or hyperbolic tangent (tanh)), we produce a single output feature map. Repeat this procedure for the rest of weight filters, we obtain the whole output feature map. Zero padding is adopted at the border to ensure the output feature maps have the same size as input. The expression of the CONV layer operation is shown in Equation (\ref{eqn:conv}). 
\vspace{-0.2cm}
\begin{equation}\label{eqn:conv}
\mathbf{y_{c'}}=f(\sum_{c=1}^{C}\mathbf{x}_{c} * \mathbf{w}_{c,c'}+\mathbf{b}_{c'})
\end{equation}
there are $C$ input feature maps $(\mathbf{x}_1; \mathbf{x}_2; \mathbf{x}_3; ...; \mathbf{x}_C)$ and $C'$ output feature maps $(\mathbf{y}_1; \mathbf{y}_2; \mathbf{y}_3; ...; \mathbf{y}_{C'})$, respectively. The weight parameters of this CONV layer ${\bf{W}}$ has the shape of $[filter\_height,~filter\\ \_width ,~ in\_channels,~  out\_channels]$, denoted as ${\bf{W}}\in\mathbb{R}^{r \times r \times C \times C'}$. $f$ is the activation function and ${\bf{b}}$ is the bias as the same size as output feature maps.

\subsection{Pooling layer and other types of layers}


In YOLO network structure, pooling layers downsample each input feature map  by passing it through a $2 \times 2$ max window (pool) with a stride of 2, resulting in no overlapped regions. Pooling layers reduce the data dimensions and mitigate over-fitting issues. Max pooling is the dominant type of pooling strategy in state-of-the-art DCNNs due to its higher overall accuracy and convergence speed \cite{chen2017eyeriss}.
Batch normalization (BN) normalizes the variance and mean of the features across examples in each mini-batch~\cite{ioffe2017batch}, to avoid the gradient vanishing or explosion problems~\cite{ioffe2015batch}.

\section{Compressed Convolution Layers }

\subsection{Block-Circulant Matrices}

We can reducing weight storage by replacing the original weight matrix with one or multiple blocks of circulant matrices, where each row/column is the cyclic reformulation of the others, as shown in Equantion~(\ref{eqn:circulant_matrix}).
\vspace{-0.2cm}
\begin{equation}
\small
\label{eqn:circulant_matrix}
\mathbf{\bf{W}}_{ij}=
\begin{bmatrix}
 w_{1}      & w_{L_b}      & \cdots & w_{3}  & w_{2}      \\
 w_{2}      & w_{1}      & \cdots  & w_{4} & w_{3}      \\
 \vdots & \vdots & \ddots & \vdots \\
 w_{L_{b-1}}      & w_{L_{b-2}}      & \cdots & w_{1} & w_{L_b}      \\
 w_{L_b}      & w_{L_{b-1}}      & \cdots & w_{2}  & w_{1}      \\
\end{bmatrix},
\end{equation}
where $L_b$ represents the row/column size of each structured matrix (or block size, FFT size).
\subsection{Block-Circulant Matrices-Based CONV}
\label{sec:structured}


CirCNN~\cite{ding2017circnn} and C-LSTM~\cite{Wang2018clstm} have a detailed discussion of the inference algorithm for block circulant matrix-based DNNs.
The theoretical foundation is derived in \cite{zhao2017theoretical}, which shows that the ``effectiveness'' of block circulant matrix-based DNNs is the same with DNNs without compression. However, CirCNN and C-LSTM do not thoroughly discuss convolutional (CONV) layers which are the major computation in state-of-the-art DNNs. In this section, we present the detailed formulation of block circulant matrix-based computation for CONV layers, which are the major computation part of the tiny YOLO algorithm. 

Given an input and weight filters (tensor), in a 2-D convolution operation, we slide each filter over all spatial locations of the input, multiply the corresponding entries of the input and filter, and accumulate the intermediate product values. The
result is the output of the 2-D convolution. It is well-known that the multiplication-and-accumulation (MAC)  operation dominates the overall convolution computation, and will be our focus of acceleration.

\begin{figure}[t]
\centering
\includegraphics[width = 0.75\columnwidth]{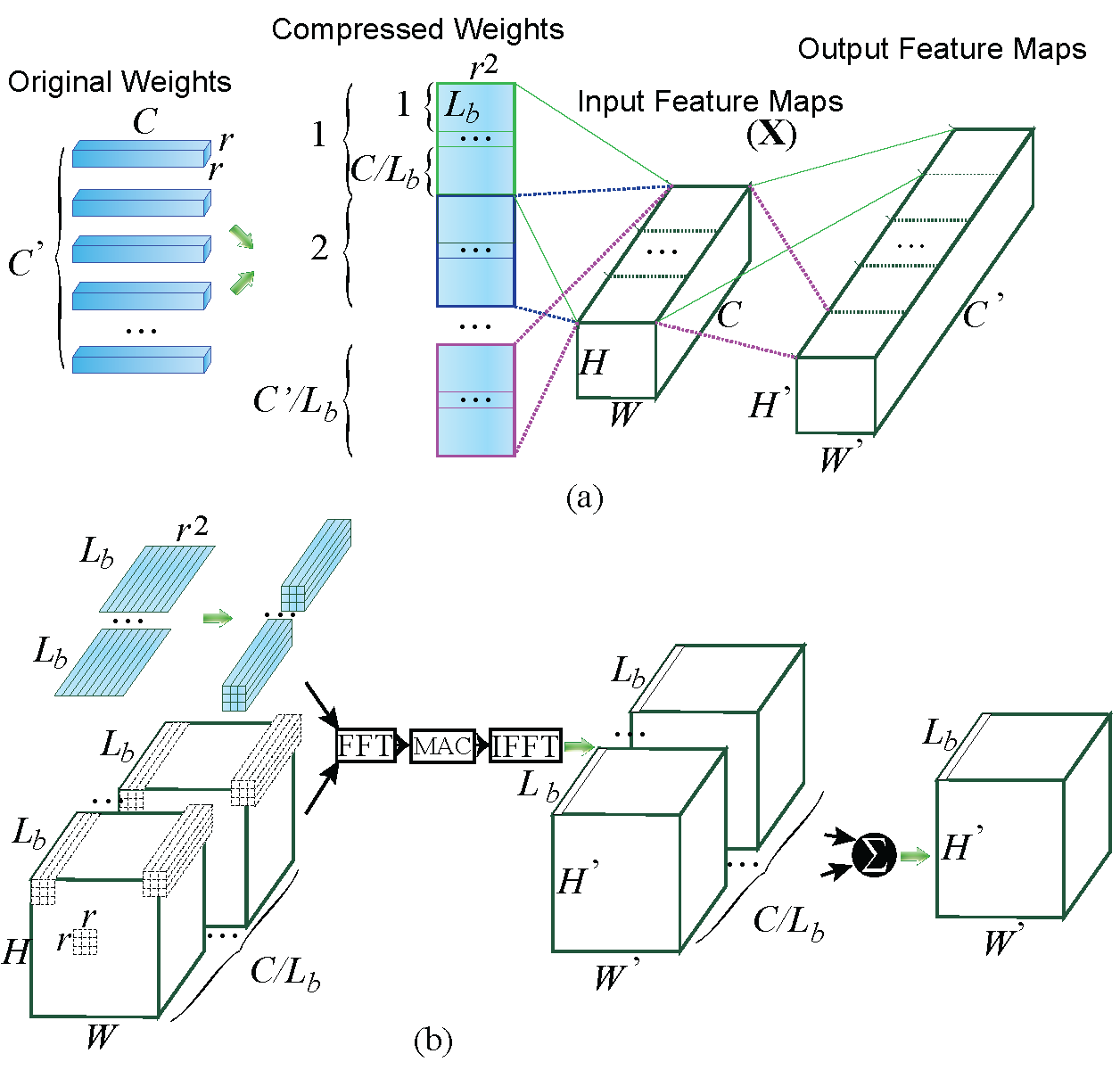}
\caption{A block circulant matrix-based CONV layer.}
\label{fig:CONV}
\vspace{-0.1in}
\end{figure}

Each 4-D weight tensor ${\bf{W}}$ has the shape of 
${\bf{W}}\in\mathbb{R}^{r \times r \times C \times C'}$. Using block circulant matrix, we compress the input channels $C$ and output channels $C'$ plane of the 4-D weight tensor. According to the \emph{circulant convolution theorem} \cite{pan2012structured,smith2007mathematics}, instead of directly performing the matrix-vector multiplication, we could use the FFT-based fast multiplication method. In each block circulant matrix, only the first row is needed for calculation, and is termed the \emph{index vector}. The calculation of a block circulant matrix-vector multiplication $\mathbf{W}_{ij} \mathbf{x}_j$ can be performed as follows.
\vspace{-0.2cm}
\begin{equation}\label{eqn:fftifft}
\mathbf{a}=\mathbf{W}_{ij}\mathbf{x}_j=\mathbf{w}_{ij}\circledast \mathbf{x}_j=\text{IFFT}\big(\text{FFT}(\mathbf{w}_{ij})\circ\text{FFT}(\mathbf{x}_j)\big)
\end{equation}
where `$ \circledast$ ' denotes circular convolution, and $\circ$ represents element-wise multiplication.
After the weight compression, the shape of the weight tensor becomes $r \times r \times C \times C'/L_b$. For better illustration, we stretch the compressed weight tensor from the 2-D $r \times r$ matrix to a 1-D $r^2$ vector. This procedure is shown in Fig.~\ref{fig:CONV} (a). The outer-level brackets indicate the IDs of the output channels for weight tensor (i.e., 1, 2, ..., $C'/L_b$). The inner-level brackets show the IDs of index vectors for input channels (i.e., 1, 2, ..., $C/L_b$), where each vector with length $L_b$ corresponds to a block circulant matrix.

Fig.~\ref{fig:CONV} (b) illustrates the block circulant matrix-based CONV operation. We slide each block of FFT results of the weight kernel FFT$(\mathbf{w}_{ij})$ (marked as blue bars) over all spatial locations of the input feature maps FFT$(\mathbf{x}_{j})$ of FFT results  with zero padding (marked as white bars and dotted lines).
We then multiply the corresponding FFT results of the input feature map FFT$(\mathbf{x}_{j})$ and weight kernel FFT$(\mathbf{w}_{ij})$. The accumulation of the $r^2$ multiplication results will be sent to IFFT computing module and become the output of the block circulant matrix-based CONV operation. 
The forward propagation process in the CONV layer is shown in 
Fig.~\ref{fig:CONV} (a). Please note that each weight kernel has $C/L_b$ blocks. After finding the CONV kernel result, 
we sum up all the intermediate matrices and output a single matrix. This becomes one channel at the output of CONV layer.  
In other words, we use the first $C$-channel of the weight kernel to compute the first output channel of CONV layer, and so on. In addition, each output channel has its own batch normalization and bias, which will be calculated in the end.

\section{The REQ-YOLO Framework}

\subsection{Heterogeneous Weight Quantization}

The previous works on weight quantization~\cite{lin2015neural,wu2016mobile} has demonstrated the effectiveness of various quantization techniques applied to DNN models including fixed bit-length, ternary and even binary weight representations. Weight quantization can simultaneously reduce the DNN model size, computation and memory access intensity \cite{courbariaux2015binaryconnect}. Some the other prior works have investigated the combination of equal-distance weight quantization and other mode compression techniques such as weight pruning~\cite{han2015deep,han2017ese}.

\begin{figure}[t]
    \centering
    \includegraphics[width=0.65\columnwidth]{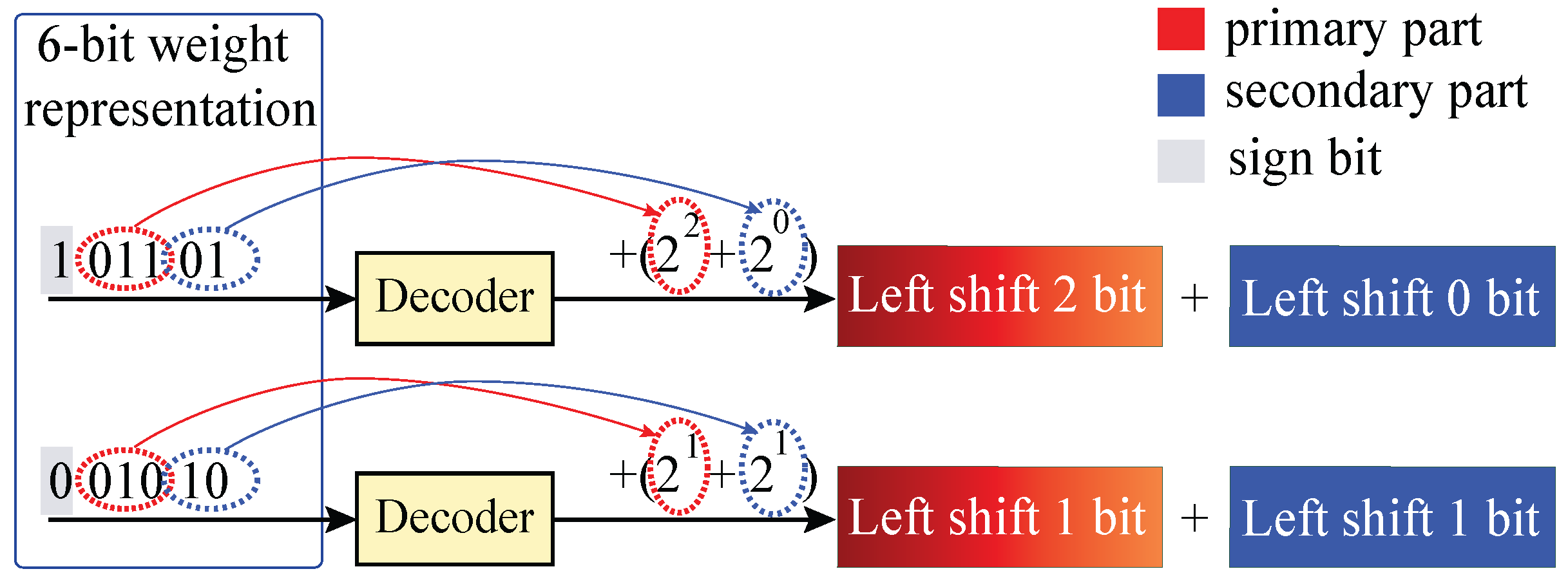} 
    \vskip -0.5em
    \caption{An illustration of the 6-bit weight representation using the mixed powers-of-two quantization.}
    \label{fig:powers_of_two} 
 \end{figure}

Equal-distance quantization~\cite{courbariaux2015binaryconnect}, to some extent, facilitates efficient hardware implementations while maintaining accuracy requirement, whereas the power consumption and hardware resource utilization of the involved multiplications is still high. On the other hand, the powers-of-two quantization technique is extremely hardware efficient by using binary bit shift-based multiplication, however, suffering non-negligible accuracy degradation due to the highly unevenly spaced scales. 
To overcome the accuracy degradation problem and maintain low power consumption and resource utilization, we propose a heterogeneous weight quantization technique, i.e., we (i) adopt the equal-distance quantization for some CONV layers and (ii) we use the mixed powers-of-two-based quantization for other CONV layers. Please note that the quantization technique is identical inside each CONV layer. The mixed powers-of-two-based weight representation consists of a sign bit part and a magnitude bits part. The magnitude bits part is the combination of a primary powers-of-two and a secondary powers-of-two part. It not only enhances the model accuracy by mitigating the uneven data scaling problem but also facilitates efficient hardware implementations. Fig.~\ref{fig:powers_of_two} illustrates a 6-bit weight representation using the mixed powers-of-two quantization method. 1 bit is for representing the sign bit, and 5-bit are for magnitude bits (in which the first 3-bit are primary; the last 2-bit are secondary). The primary and secondary part of the weight "101101" are decoded as "011" and "01", respectively. Therefore, when multiplying an input value by weight "101101", we shift the input left 2 bit and 0 bit, respectively, and sum the two products up.

\subsection{ADMM for Weight Quantization}
\label{admm}

\begin{figure} [t]
     \centering
     \includegraphics[width=0.75\columnwidth]{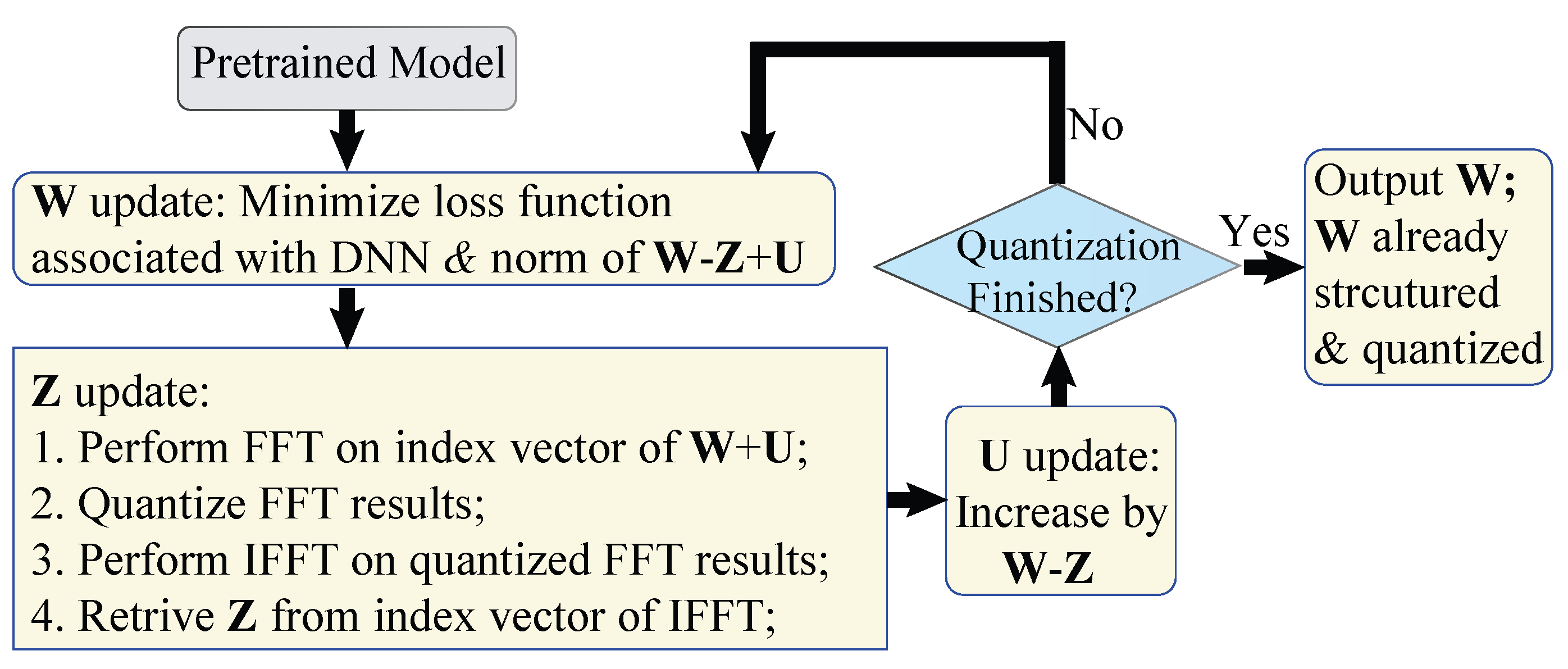}    
     \caption{The overall procedure of ADMM-based weight quantization on FFT results.}
     \label{fig:ADMM_fft}
\vspace{-0.1em}
\end{figure}

In the hardware implementation, for each block circulant matrix ${\bf{W}}_{ij}$, we actually store the FFT result $\text{FFT}({\bf{w}}_{ij})$ instead of the index vector ${\bf{w}}_{ij}$~\cite{ding2017circnn}. 
However, it is not straightforward to directly apply quantization on the FFT results $\text{FFT}({\bf{w}}_{ij})$'s because of the difficulty of impact evaluations. This is a major limitation of the prior work~\cite{han2015deep,han2017ese,ding2017circnn,Wang2018clstm}, which would be further exacerbated because both real and imaginary parts of FFT results need to be stored.

To overcome this limitation, in this section we incorporate ADMM with FFT/IFFT and use it for the heterogeneous weight quantization to directly quantize the FFT results $\text{FFT}({\bf{w}}_{ij})$'s, which can achieve higher compression ratio and lower accuracy degradation compared with prior works. This novel method effectively leverages the flexibility in ADMM. In a nutshell, we propose to \emph{perform quantization in the frequency (FFT) domain and perform weight mapping in the weight domain}. 
Details are described as follows, as also shown in Fig.~\ref{fig:ADMM_fft}.

  Consider the quantization problem as an optimization problem $\min_{{\bf{x}}} f(\bf{x})$ with combinatorial constraints. This problem is difficult to solve directly using optimization tools. Through the application of ADMM~\cite{boyd2011distributed,jin2017deep}, the original quantization problem is decomposed into two subproblems, which will be iteratively solved until convergence. The first subproblem is $\min_{{\bf{x}}} f({\bf{x}})+q_1(\bf{x})$ where $q_1(\bf{x})$ is a differentiatede, quadratic term. This subproblem does not have combinatorial constraints and can be solved using traditional optimization method, e.g., stochastic gradient descent for DNN training.   The second subproblem is: $\min_{{\bf{x}}} g({\bf{x}})+q_2(\bf{x})$, where $g({\bf{x}})$ corresponds to the original combinatorial constraints and $q_2(\bf{x})$ is another quadratic term. For special types of combinatorial constraints, including block circulant matrices, quantization, etc., the second subproblem can be optimally and analytically solved, as we will see in the following discussions.

In the tiny YOLO network, the weights in the $l^{th}$ layer is denoted by ${\bf{W}}_{l}$. The loss function is represented by $f \big( \{{\bf{W}}_{l}\}_{l=1}^N\big)$. 
Assume a weight sub-matrix $({\bf{W}}_{l})_{ij}\in\mathbb{R}^{L_b\times L_b}$ is mapped to a block circulant matrix. Directly training the network in the structured format will incur a large number of equality constraints (to maintain the structure). This makes the training problem inefficient to solve using the conventional stochastic gradient descent. On the other hand, ADMM can be utilized to efficiently solve this problem, and a large number of equality constraints can be avoided.

We introduce auxiliary variables ${\bf{Z}}_{l}$ and ${\bf{U}}_{l}$, which have the same dimensionality as ${\bf{W}}_{l}$. Through the application of ADMM\footnote{The details of the ADMM algorithm are discussed in \cite{boyd2011distributed}. We omit the details because of space limitation.}, the original structured training problem can be decomposed into two subproblems, which are iteratively solved until convergence. In each iteration $k$, the first subproblem is

\vspace{-0.5cm}
\begin{equation}
\label{4}
 \underset{ \{{\bf{W}}_{l}\}}{\text{minimize}}
\ \ \ f \big( \{{\bf{W}}_{l} \}_{l=1}^N\big)+\sum_{l=1}^{N} \frac{\rho_{l}}{2}  \| {\bf{W}}_{l}-{\bf{Z}}_{l}^{k}+{\bf{U}}_{l}^{k} \|_{F}^{2}, \\
\end{equation}

where ${\bf{U}}_{l}^{k}$ is the dual variable updated in each iteration, ${\bf{U}}_{l}^{k}:={\bf{U}}_{l}^{k-1}+{\bf{W}}_{l}^{k}-{\bf{Z}}_{l}^{k}$. In the objective function of (\ref{4}), the first term is the differentiable loss function, and the second quadratic term is differentiable and convex. Thus, this subproblem can be solved by stochastic gradient descent and the complexity is the same as training the original DNN. A large number of constraints are avoided here. The result of the first subproblem is denoted by ${\bf{W}}_{l}^{k+1}$.

  The second subproblem, on the other hand, is to quantize ${\bf{W}}_{l}^{k+1}+{\bf{U}}_{l}^{k}$ in the frequency domain, and the result of the second subproblem is denoted by ${\bf{Z}}_{l}^{k+1}$. For a matrix $({\bf{W}}_{l}^{k+1}+{\bf{U}}_{l}^{k})_{ij}$ for frequency-domain quantization, we first perform FFT on the index vector. Then we perform quantization on the FFT results. For the equal-distance quantization, we constrain them on a set of quantization levels $\alpha \times \{-(\frac{M}{2}-1), ..., -1, 0, 1, 2, ..., \frac{M}{2}-1\}$ associated with a layerwise coefficient $\alpha$, where $M$ is the  predefined number of quantization levels; for the mixed powers-of-two quantization, we constraint them to $\alpha \times \{ 0, \pm 2^0,  \pm 2^1, \pm 2^2, ..., \pm 2^{M_1}\}$ $\bigcup$ $ \{0, \pm 2^0,  \pm 2^1, \pm 2^2, ..., \pm 2^{M_2}\}$, where $M_1$ and $M_2$ is the total number of bits in the primary and secondary part, respectively.

This step is simply mapping each FFT value to the nearest quantization level. The quantization levels are determined by the (i) range of FFT results of index vector, and (ii) the predefined number $S$ of quantization levels. The coefficient $\alpha$ may be different for different layers, which will not increase hardware implementation complexity because $\alpha$ will be stored along with the FFT results after quantization. Finally, as the key step, we perform IFFT on the quantized FFT results, and the restored vector becomes the index vector of $({\bf{Z}}_{l}^{k+1})_{ij}$. We then retrieve block-circulant matrix ${\bf{Z}}_{l}^{k+1}$ from the index vector after IFFT.

We have proved that the above frequency-domain quantization procedure is the optimal, analytical solution of the second subproblem. Because of the symmetric property of quantization above and below 0, the restored index vector of $({\bf{Z}}_{l}^{k+1})_{ij}$ will still be a real-valued vector. Besides, the frequency-domain quantization procedure is applied after the block circulant matrix training of DNNs, and the circulant structure will be maintained in quantization. This is because we restore a single index vector for $({\bf{Z}}_{l}^{k+1})_{ij}$ and thus $({\bf{Z}}_{l}^{k+1})_{ij}$ will maintain the imposed circulant structure. After the convergence of ADMM, the solution ${\bf{W}}_l$ meets the two requirements: (i) the block-circulant structure, and (ii) the FFT results are quantized.

\section{Hardware Implementation}


In this section, we implement the YOLO-based object detection on FPGAs. In order to achieve both low-power and high-performance, the proposed REQ-YOLO framework ensures that the limited FPGA on-chip BRAM has enough capacity to load the weight parameters from the host memory due to the following reasons: (i) the regularity of the block circulant matrices introduces no additional storage such as weight indices after compression in ESE~\cite{han2017ese}; (ii) we use the heterogeneous weight quantization using ADMM considering hardware resource, further reducing the weight storage and exploiting the hardware resource while satisfying the accuracy requirement.
The extra communication overhead caused by accessing FPGA off-chip DDR for common designs~\cite{han2017ese, nakahara2018lightweight} can be eliminated.

\subsection{FPGA Resource-Aware Design Flow}
\label{sec:resouce-aware}


The resource usage model including Look-up tables (LUTs), DSP blocks, and BRAM of an FPGA implementation can be estimated using analytical models. According to our design, there are two types of PEs: DSP-based PE for equal distance quantization and shift-based PE for mixed powers-of-two quantization. In the convolution operation, suppose the DSP resource for DSP-based and shift-based PE are $\Delta DSP_{D}$ and $\Delta DSP_{S}$, respectively, and the LUT resouce for DSP-based and shift-based PE are $\Delta LUT_{D}$ and $\Delta LUT_{S}$, respectively.
The models of \# DSP, \# LUT, and \# BRAM are shown as follows,
\begin{gather}
\label{eqn:dsp}
\# DSP =  \Delta DSP_{D} \times \#CONV_{D}+\Delta DSP_{S} \times \#CONV_{D}\\
\label{eqn:lut}
\# LUT =  \Delta LUT_{D} \times \#CONV_{L}+\Delta LUT_{S} \times \#CONV_{L}\\
\label{eqn:bram}
\# BRAM= max\{\frac{Model~size}{BRAM~size},\frac{Onchip~bandwidth}{BRAM~bandwidth}\}
\end{gather}
where $\#CONV_{D}$, $\#CONV_{L}$ are the number of CONV operations for DSP and LUT, respectively. Generally, in Xilinx Virtex-7 FPGA fabric, the BRAM size is 36kb and the BRAM bandwidth is 64b.

\begin{figure}[t]
\centering
\includegraphics[width = 0.8\columnwidth]{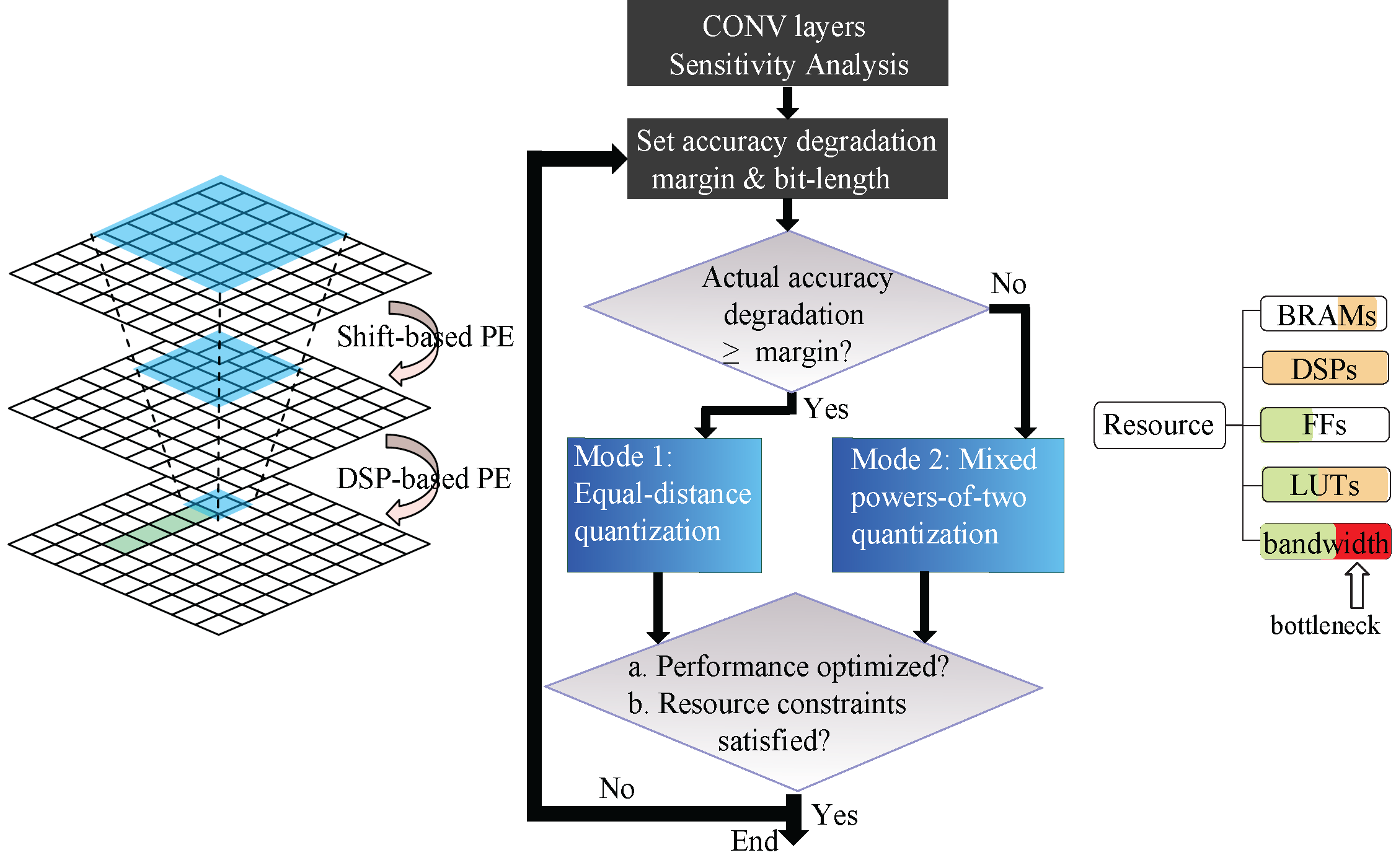}
\caption{The mode selection rule of the resource-aware design.}
\label{fig:Coop}
\end{figure}

Indeed, replacing multiplications with bit shift operations significantly reduces the usage of DSP blocks, resulting in much less power assumption. However, the DSP resource will be wasted, causing utilization overhead on LUTs since LUT is the basic building block in implementing the logic function of bit shift operations. To fully exploit the limited FPGA resource for both LUTs and DSP blocks, we propose to adopt both the equal-distance quantization and the mixed powers-of-two-based quantization techniques for hardware implementation. More specifically, for each CONV layer, we select either equal-distance or mixed powers-of-two as the quantization method. Please note that the quantization method inside a CONV layer is identical. 

The selection rule is shown in Fig.~\ref{fig:Coop}. The design objectives are higher performance and energy efficiency satisfying the accuracy requirement. We first conduct the sensitivity analysis for each CONV layer regarding the two quantization methods (mode 1 and mode 2) and we set the initial margin for the overall degradation of the prediction accuracy and initial bit-length for weight representation. In order to reduce the accuracy degradation as much as possible, our priority choice is equal-distance quantization in those CONV layers which sensitivity are beyond the pre-set margin value since we can use DSPs for multiplication operations to enhance the accuracy. More specifically, we choose mode 2 if the actual accuracy degradation of the YOLO network is smaller than the margin value, otherwise we select mode 1. The margin value will further be refined until the performance is optimized and resource constraints (i.e., DSPs and bandwidth) are satisfied. Overall, the DSPs and LUTs usage is not the bottleneck in our design, which is different from traditional fixed-bit length weight quantization and our design is bound by bandwidth only.

\subsection{Overall Hardware Architecture}

 \begin{figure}[t]
    \centering
    \includegraphics[width=0.45\columnwidth]{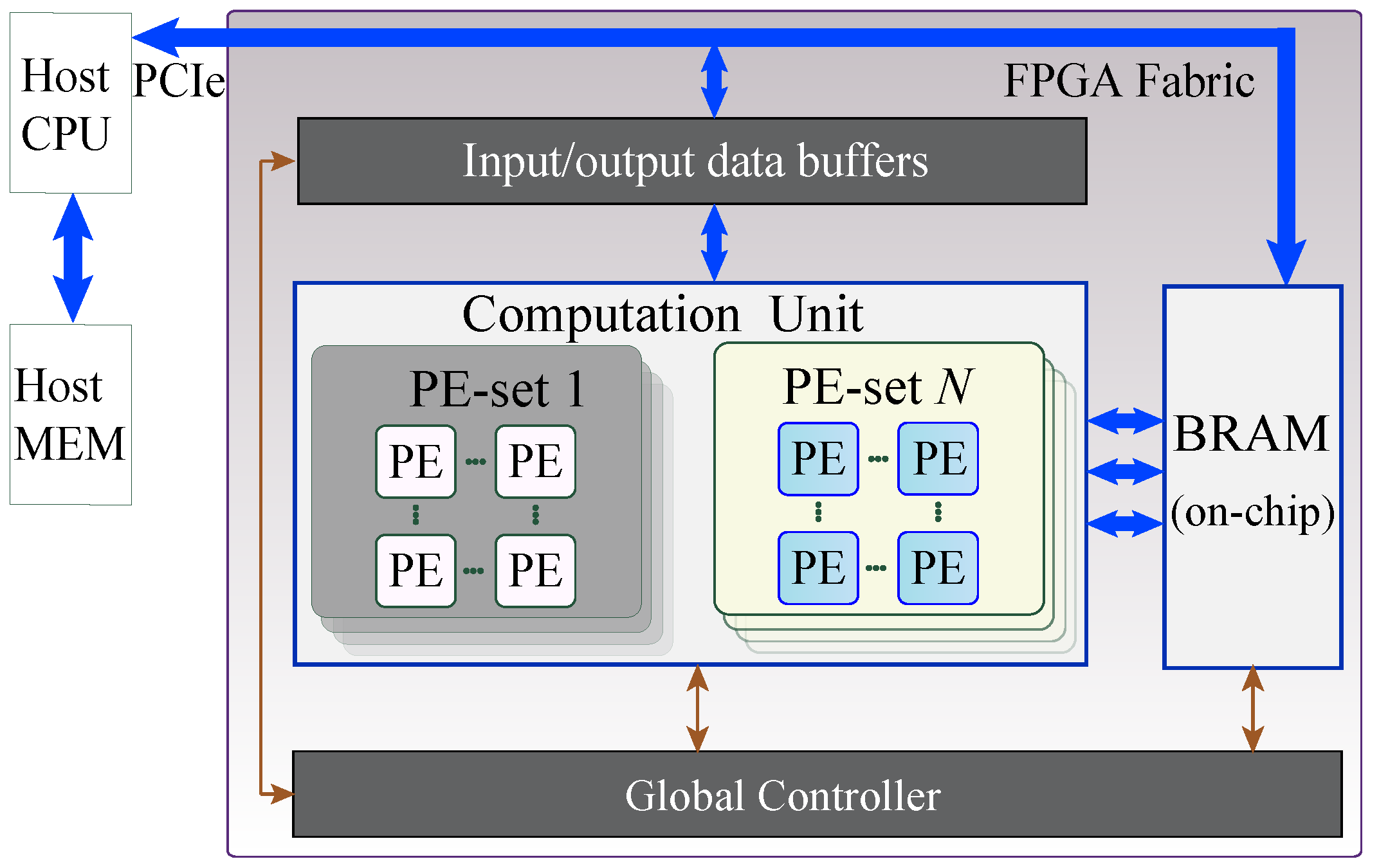} 
    \vskip -0.5em
    \caption{The overall hardware architecture on FPGA.}     
    \label{fig:sys_overall}
 \end{figure}

Our proposed accelerator design on FPGA is composed of a computation unit, on-chip memories/BRAM, and datapath control logic. Our design does not access the FPGA off-chip memories. The FFT results of the pre-trained network model (CONV weight filters) is loaded to FPGA BRAM from the Host memory via the control of Host CPU and the PCI-express (PCIe) bus. The data buffers are used to cache input images, intermediate results and layer outputs from the previous stage slice by slice, to make preparation of the PE operation. The PEs inside of the computation unit are a collection of basic computing units that execute multiplications and additions (MACs), and other functions such as normalization, etc. The global controller orchestrates the computing flow and data flow on the FPGA fabric.

\subsection{PE Design}

\begin{figure}[b]
    \centering
    \includegraphics[width=0.9\columnwidth]{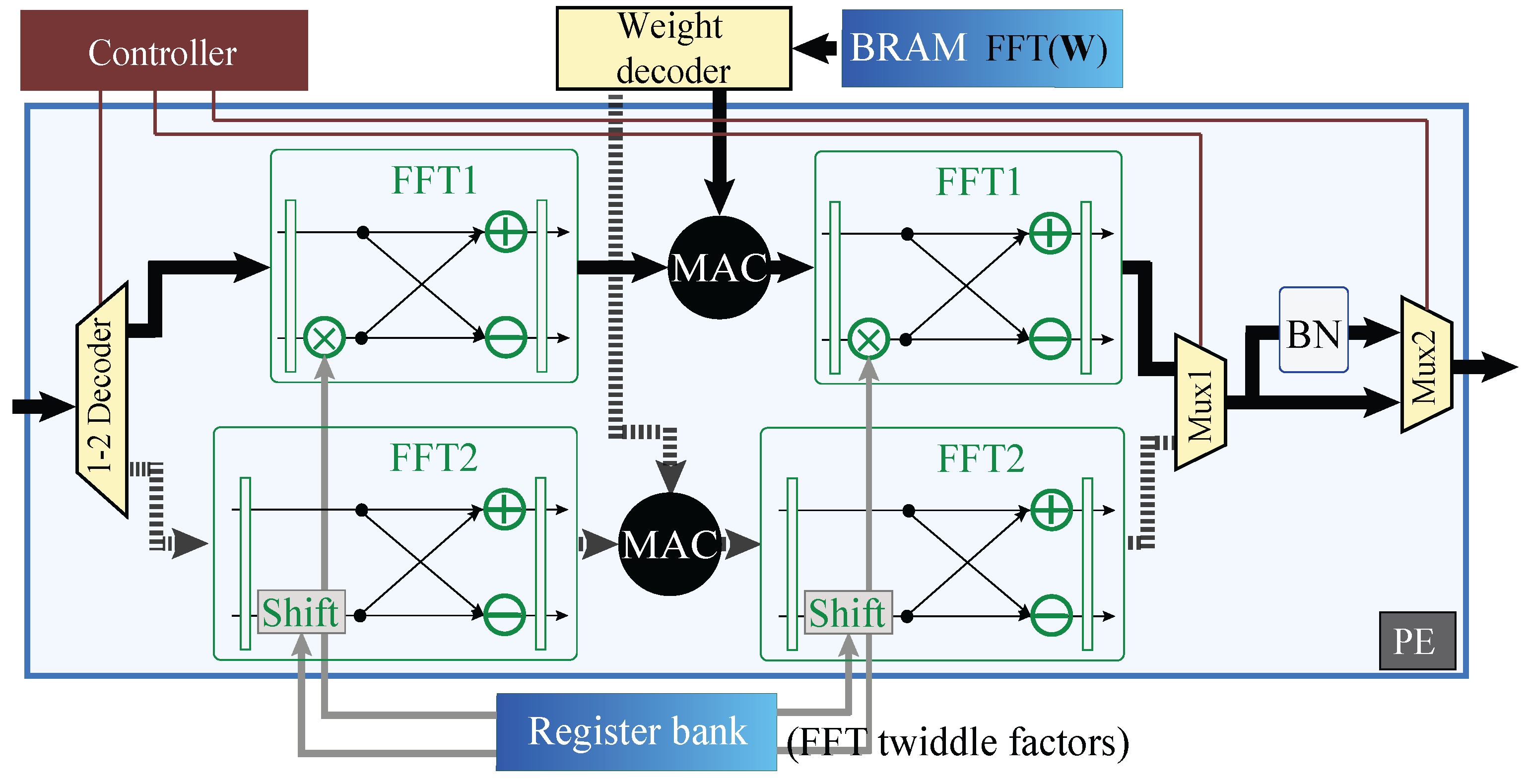} 
    \vskip -0.5em
    \caption{The PE (processing element) design.}
    \label{fig:PE} 
 \end{figure}

From Eqn.~(\ref{eqn:fftifft}), we observe that the FFT and IFFT operation are always executed in pairs. Therefore, we can combine and implement them as an FFT/IFFT kernel. 
An $N$-point IFFT calculation can be implemented using an $N$-point FFT in addition of a division operation (i.e., $\div N$) and two conjugations. There are $N$ multipliers between FFT and IFFT, which is responsible for multiplying the intermediate results of FFT and weight values stored in BRAM. 
The PE is designed mainly to execute the most resource-consuming operation, i.e., matrix-vector multiplication, which will be implemented using the element-wise  ``FFT$\rightarrow$MAC$\rightarrow$IFFT'' calculation according to Equation~\ref{eqn:fftifft} (the key for the implementation with limited hardware resources). As shown in Fig.~\ref{fig:PE}, the proposed PE architecture consists of a register bank, a controller, a weight decoder, a mode decoder, 2 multiplexers, and 2 FFT/IFFT kernels. 

The register bank stores the FFT twiddle factors which will be loaded to the FFT operator. The weight decoder prepares the desired weight parameters format for further calculation. The mode decoder and MUX 1 work simultaneously to select the operating mode (i.e., mode 1 for equal-distance quantization and mode 2 for mixed powers-of-two quantization), under the control of the PE controller. Mux 2 is used to select the batch normalization (BN) operation depending on the layer structure. The MAC unit multiplies and accumulates the input feature maps and the pre-calculated FFT result of convolution kernel weights decoded from BRAM blocks.
The two operating modes are marked in red dashed boxes, where mode 1 performs FFT/IFFT computations using multiplication-based FFT butterfly unit~\cite{cooley1965algorithm} along with a MAC unit, while mode 2 performs the FFT/IFFT computations using binary bit shifts and additions. Additionally, the MAC operation can be replaced by shift and addition in model 2.

\subsection{Convolution Dataflow and Pipelining}

 \begin{figure}[t]
    \centering
    \includegraphics[width=0.75\columnwidth]{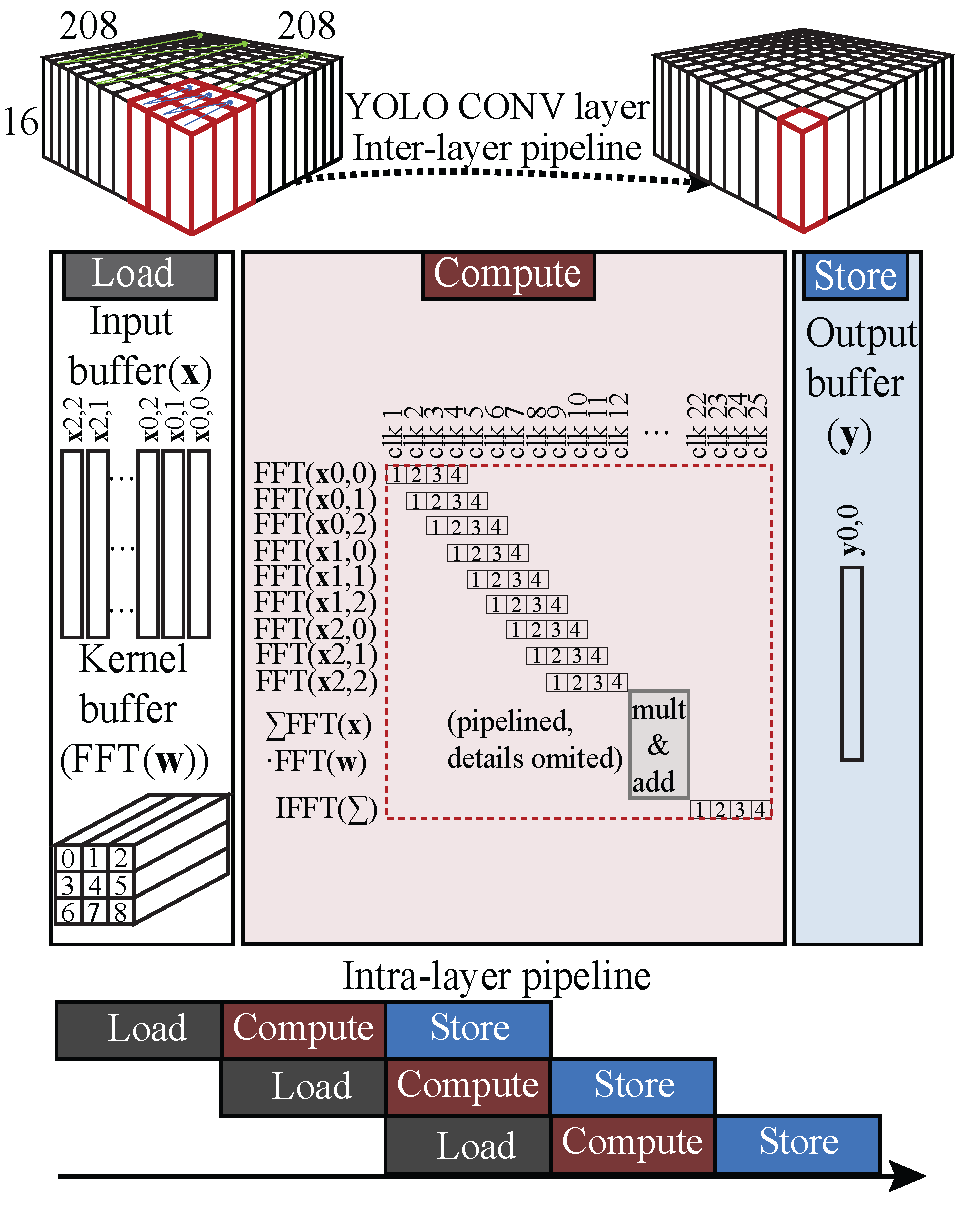}
    \vskip -0.5em
    \caption{An illustration of the CONV operation data flow in the $2^{nd}$ CONV layer. }
    \label{fig:dataflow} \vspace{-0.2cm}
 \end{figure}

The dataflow of input pixels from input buffers to computation unit to output bufferers is shown in Fig. \ref{fig:dataflow}. 
Taking advantage of the compressed but regular YOLO network structures, 
we apply inter-level and intra-level pipelining in the basic computation unit
as shown in Fig. \ref{fig:dataflow}. 
In \emph{intra-level pipelining}, there are three separate stages, i.e., load, compute, and store. This pipelining scheme generates higher level of parallelism and therefore leads to higher performance and throughput. In the \emph{inter-level pipelining}, each pipeline stage corresponds to one level in the computation unit. There are several stages in the FFT operation depending on the input size, i.e., an $N$-point FFT uses $N/2$ butterfly units for each stage and has a total of $r=$log$_2^N$ stages. 

We use an input size of $208 \times 208 \times 16$ and  weight kernel size $3 \times 3 \times 16$ using 16-point FFT (4 stages) as shown in Fig. \ref{fig:dataflow}, to demonstrate the CONV dataflow in the $2^{th}$ CONV layer of the tiny YOLO network structure. Both inter-level and intra-level pipelining techniques are adopted. The input pixels are loaded to input buffers followed by a sequence corresponding to the spatial relationship with kernel window. The first input data needed for CONV operation is marked as red bars with the same input size of 16-point FFT. The PE accepts each input vector (red bar) each per clock cycle, computes the ``FFT$\rightarrow$MAC$\rightarrow$IFFT'' operation and the result is stored in the output buffer.

\subsection{Design Optimization}

In YOLO, the CONV operation, performed by PEs, is the most resource-intensive arithmetic.
Therefore, from the perspective of computation, the hardware design optimization targets at PE size/number. Through reducing PE size/number, we can achieve less power and area consumption, leading to more available on-chip resource and more parallelism. 
From the communication perspective, the cost of moving data from one physical location to another on FPGAs, named communication cost, can dominate the computational energy and our design. Communication cost consists of accessing memory of weight parameters and intermediate results, and moving data bits over interconnect wires between PEs. Therefore, we can optimize the required computation and communication cost by reducing PE size/number including LUTs and DSPs, and memory access.

\subsubsection{DSP Usage Optimization}
Reducing  the  number  of  multiplications  will  be critical to the overall hardware design optimization. In the FFT operation, the multipliers of the Radix-2 FFT butterflies with twiddle factor 1 and -1 can be eliminated, and the multiplier of those butterflies with twiddle factor $j$ and $-j$ can be replaced with conjugation operators. In the dot product stage, the two inputs of dot product are both from FFT operators, which are conjugate symmetric. And the dot product results of such conjugate symmetric inputs are also conjugate symmetric. Therefore, in $\big($FFT($\mathbf{x}_j$) $\circ$ FFT($\mathbf{w}_{ij}$)$\big)$ with input size of FFT $N$, the last $N/2-1$ dot product outputs can be obtained using conjugation operations from their corresponding symmetric points. And the amount of $N/2-1$ multipliers can be eliminated.

 
The DSP48E1 block in modern Xilinx FPGAs generally consists of three sub-blocks: pre-adder, multiplier, and ALU. These hard blocks directly implement commonly used functions in silicon, therefore consuming much less power and area, and operating at a higher clock frequency than the same implementations in logic. For these hard blocks with constrained resource, resource sharing could be applied. Generally, non-overlapping MAC operations are scheduled using the combination of {pre-adder, multiplier} blocks or {ALU, multiplier} blocks based on the function itself and bit-length of operands. 

In order to take full advantage of the limited DSP resource and achieve more design parallelism, we further optimize the proposed design using low-bit DSP sharing. More specifically, we can divide each sub-block into smaller slices, in which the internal carry propagation between slices is segregated to guarantee independence for all slices. In other word, we can group and feed several non-overlapping operands into one of the inputs of a DSP sub-block. For example, the ALU unit in DSP48E1 block can be divided into six 8-bit smaller slices with carry out signal for 8-bit computation.

  \subsubsection{Reducing Weight Memory Accesses.}


The input feature map $\mathbf{x}_i (i\in (1,...,C))$ is real value~\cite{everingham2010pascal} and all the weight parameters $\mathbf{W}$ are real-valued. According to~\cite{salehi2013pipelined,chang2003efficient}, the FFT or IFFT result is mirrored (have the property of complex conjugated symmetry) when its inputs are real-valued. Therefore, for an FFT/IFFT with $N$-point inputs, we only need to store $\frac{N}{2}+1$ of the results instead of $N$ results into the BRAM, thereby reducing communication energy.

\subsection{Design Space Exploration}


\begin{algorithm}[t]
\small
 \KwIn{$\#CONV_D, \#CONV_L, \Delta DSP_D, \Delta DSP_S, \Delta LUT_D,\Delta LUT_S$, BRAM size and bandwidth, YOLO model Size, and onchip bandwidth.}
 \KwOut{bit-length $b$, mode type of $i_{th}$ layer ${\bf M}_i (i \in (0,...,8))$.}
 Analyze the sensitivity of all the CONV layers;\\
 Set initial $b$ \& accuracy degradation margin $\Delta ACC_{m}$;\\
 \For{$i \gets 0$ until $n\_layers$}
        {${\bf M}_i \gets$ 2;\\
   \While {resource \& performance not optimized}
     {Change $b~or~ \Delta ACC_{m}$;\\
         \If{actual accuracy loss $\Delta ACC_{act} \geq \Delta ACC_{m}$}
         {${\bf M}_i \gets$ 1;} 
        Calculate resource usage \#DSP, \#LUT, and \#BRAM using 
        Equation~(\ref{eqn:dsp},~\ref{eqn:lut},~\ref{eqn:bram});}
    }
 \Return{$b,\mathbf{M}$.}
 \caption{Pseudo-code for resource-aware exploration}
 \label{alg:1}
\end{algorithm}

To prototype and explore the hardware architecture of the proposed REQ-YOLO framework, we use Xilinx SDx 2017.1 as the commercial synthesis backend
to synthesize the C/C++ based YOLO LSTM design. We feed the well-trained inference models of YOLO into the automatic synthesis backend~\cite{Liang:2012:JECE,wang2017comprehensive,wang2017flexcl}. A bit-length of data quantization and mode selection for each CONV layer are generated to illustrate the computation flow as shown in Algorithm~\ref{alg:1}. The operators in each graph are scheduled to compose the intra-layer or inter-layer pipeline under the design constraints, to maximally achieve full throughput and performance. At last, a code generator receives the scheduling results and generates the final C/C++ based codes, which can be fed into the commercial HLS tool for FPGA implementation.

\section{Evaluation Results}
%
\subsection{Training of Tiny YOLO}
We adopt the state-of-the-art object detection algorithm-tiny YOLO based on yolov2 tiny~\cite{darkflow} as the target DNN and evaluate it on both the PASCAL VOC07+12 dataset~\cite{everingham2010pascal} and the DataDJI detection dataset captured by the DJI UAV~\cite{DJI}. We set $S$ = 13, $B$ =5. The DataDJI detection dataset has 12 labeled classes so $C_{DJI}$ = 12, while the PASCAL VOC dataset has 20 labeled classes so $C_{VOC}$ = 20. The anchor boxes sizes are pre-set by K-means clustering with $K = 5$. 
Our final prediction is a $13 \times 13 \times 37$ tensor for DataDJI dataset and a $13 \times 13 \times 45$ tensor for VOC dataset. After the last convolutional layer, 5 boxes for each grid cell will be obtained with their locations and scores. Then we first discard boxes that have detected a class with a score lower than the threshold, which is 0.6 in our experiment. After that, the Non-Maximum Suppression (NMS) algorithm will filter out remaining boxes that overlap with each other. The ideal output of YOLO is one bounding box for each object. Finally, for the predicted bounding box, we use \emph{IOU} and mean
average precision ($mAP$) as the metric to evaluate the object detection accuracy on DataDJI and VOC datasets, respectively. 

\begin{figure}[t]
    \centering
    \includegraphics[width=0.8\columnwidth]{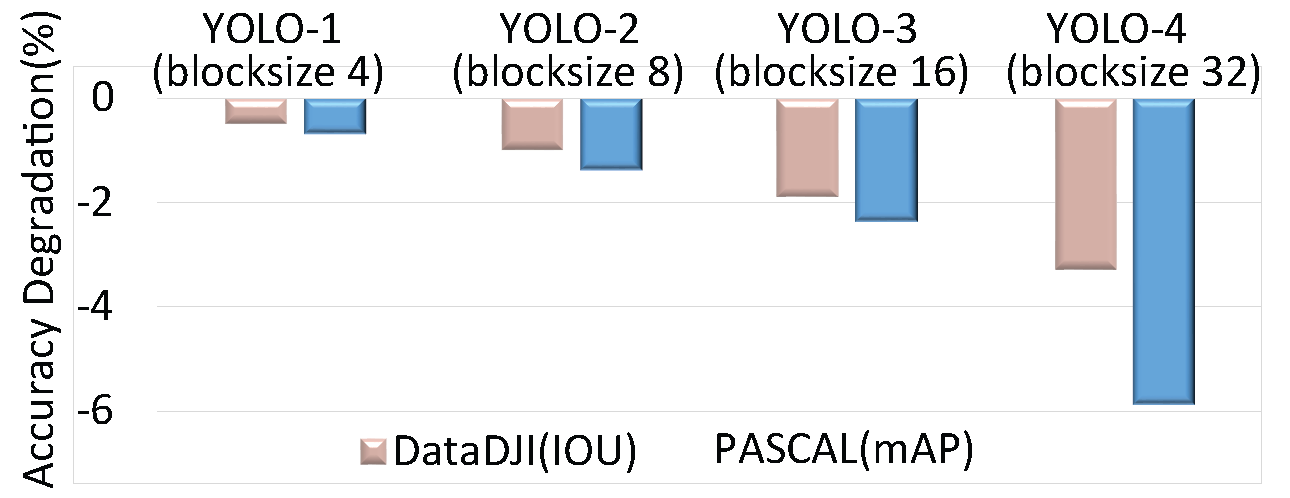} 
    \caption{Test accuracy of the tiny YOLO network using different block sizes.}
    \label{fig:acc} 
    \vspace{-0.4cm}
 \end{figure}

For both datasets, we use a mini batch size of 16. The initial learning rate is set to 0.001, and divided by a factor of 10 every 20k iterations, to guarantee convergence. The ADAM \cite{kingma2014adam} optimizer and standard data argumentation like random crops, color shifting, etc are used during training. The training results of the tiny YOLO using different block sizes are shown in Fig~\ref{fig:acc}. 
Fig.~\ref{fig:acc} shows that the block circulant matrix-based training only causes a very small accuracy degradation ($IOU$ or $mAP$) in general when the compression ratio is large. Among the four models, we select the YOLO-3 for the ADMM-based heterogeneous weight quantization and further hardware implementation, since it introduces small accuracy loss while maintaining the large compression ratio compared to baseline. 

\subsection{Accuracy after Weight Quantization}

we select the YOLO-3 model with very small accuracy degradation and large weight reduction ratio for the REQ-YOLO framework and evaluate the selected model on both the PASCAL VOC dataset and the DataDJI detection dataset. 
For different weight (FFT results) representations from 32-bit to 6-bit, the introduced additional accuracy degradation (i.e., $IOU$ for DataDJI and $mAP$ for VOC) is generally very small, i.e., 0.73\% for PASCAL, and 0.2\% for DataDJI. Reducing the weight from 32-bit floating point to 8-bit fixed point brings negligible additional degradation (0.07\%) in $IOU$ or $mAP$ while the model size can further be compressed by 4$\times$. In this way, through block circulant matrix training and ADMM-based heterogeneous quantization, we can accommodate the tiny YOLO structures to the on-chip BRAM of state-of-the-art FPGA while achieving real-time object detection, satisfying the accuracy requirement. 

\subsection{Performance and Energy Efficiency}

\begin{table*}[t]
    \centering
    \caption{Comparison of equal-distance-based quantization method and heterogeneous-based quantization method on the YOLO-3 model (block size 16).
    }
    \label{tbl:yolo}
    \resizebox{1.5\columnwidth}{!}{
        \begin{tabular}{|c|c|c|c|c|c|c|c|c|c|}
            \hline
            \multirow{2}{*}{Model}  & Layer& Comp.
            & \multicolumn{2}{|c|}{Comm.} &Bound&\multicolumn{2}{|c|}{Equal-distance-based} &\multicolumn{2}{|c|}{Heterogeneous-based}\\\cline{4-5}\cline{7-10}
            & & Size 
            & In\_size & Out\_size &Type&Latency ($\mu$ s)&Model Size&Latency ($\mu$ s)&Model Size\\\cline{1-10}
            \multirow{9}{*}{YOLO-3}&$Conv0$ &173,056        &
            519,168 & 692,224  &Comm.-bound&872.5&0.16kb&881.3&0.13kb\\ \cline{2-10}
        &    $Conv1$ &86,528         
        &692,224   &32,448   &Comm.-bound&442.2&6.75kb&443.6&5.63kb\\ \cline{2-10}
         &   $Conv2$ & 21,632       
         &32,448   &173,056  &Comm.-bound&219.3&27.0kb&216.2&22.5kb \\ \cline{2-10}
            &$Conv3 $ &21,632         
            &173,056   &86,528  &Comm.-bound&119.8&108.0kb&120.5&90.0kb \\ \cline{2-10}
           & $Conv4$ & 21,632       
           &86,528    &43,264  &Comp.-bound&117.1&432.0kb&54.5&360.0kb\\ \cline{2-10}
            &$Conv5$ &21,632          
            &43,264   &86,528   &Comp.-bound&117.9&1.69Mb&69.4&1.41Mb\\ \cline{2-10}
           & $Conv6$ &86,528         
           &86,528   &173,056   &Comp.-bound&905.1&3.38Mb&430.4&2.81Mb\\ \cline{2-10}
           & $Conv7$ &173,056         
           &173,056    &173,056 &Comp.-bound&1,832.7&286.88kb&872.7&239.06kb \\ \cline{2-10}
           &$Conv8$ &16,224          
           &173,056  &19,244   &Comp.-bound&174.3&37.88kb&93.2&26.57kb \\ \cline{2-10}
             &  Total &{\bf 621,920}  
             &-  &-   &-&{\bf 4,801.0}&{\bf 5.93Mb}&{\bf 3,183.6}&{\bf 4.95Mb} \\\hline

        \end{tabular}
    }
\end{table*}

\begin{table*}[t]
    \centering
    \caption{Comparison among different tiny YOLO implementations.}
    \label{tb1:platform}
    \resizebox{2.0\columnwidth}{!}{
         \begin{tabular}{|c|c|c|c||c|c|c|c|}
            \hline 
           \multirow{2}{*}   {Implementation} & Titan X-YOLO & Our GTX-YOLO  & Our TX2-YOLO & Virtex-YOLO&Zynq-YOLO & \textbf{Our FPGA-YOLO0} & \textbf{Our FPGA-YOLO1 }\\ 
           &~\cite{redmon2016you}&&&~\cite{ma2017hardware}&~\cite{guo2016model}&\textbf{(Equal-distance-based)}&\textbf{(Heterogeneous-based)}\\\hline
           Device Type  &Titan X&GTX 1070 GPU      &TX2 embedded GPU&Xilinx Virtex-7 485t &Zynq 7020  & ADM-7V3 FPGA  & ADM-7V3 FPGA \\\hline
            Memory &12GB GDDR5& 8GB GDDR5       &  8 GB LPDDR4 & 4.5 MB BRAM& 0.6 MB BRAM&6.6 MB BRAM & 6.6 MB BRAM  \\\hline

            Clock Freq. &1.0 GHz & 1.6 GHz       & 1.3 GHz& 0.14 GHz&0.15 GHz (Peak) & 0.2 GHz & 0.2 GHz \\\hline
            Performance (FPS)&155 & 220.8     & 28.4 &21&8 & \textbf{208.2} & \textbf{314.2} \\\hline
            Power (W) & 180 &  140      & 10.8  &-&-& \textbf{23} & \textbf{21}  \\\hline
            Energy Efficiency (FPS/W) & 0.9& 1.6       & 2.6&- & -&\textbf{9.1}& \textbf{15.0} \\\hline
               \end{tabular}   
        }
\end{table*}

\subsubsection{FPGA-platform Comparison}
\begin{figure}[t]
    \centering
    \includegraphics[width=0.6\columnwidth]{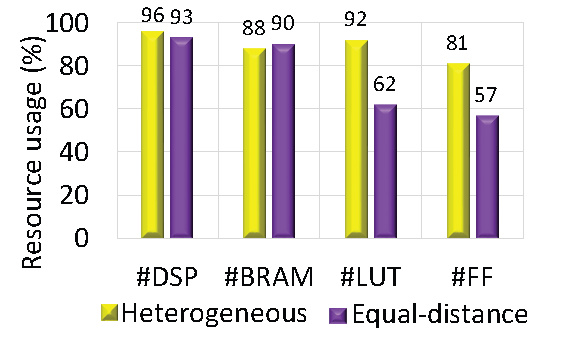} 
    \caption{Resource usage comparison of two different quantization methods.}
    \label{fig:resource} \vspace{-0.2cm}
 \end{figure}

We use the FPGA platform of Alpha Data's ADM-PCIE-7V3 for evaluating the proposed REQ-YOLO framework. The ADM-PCIE-7V3 board, comprising a Xilinx Virtex-7 (690t) FPGA and a 16GB DDR3 memory, is connected to the host machine through PCIE Gen3 $\times$ 8 I/O Interface. The host machine adopted in our experiments is a server configured with multiple Intel Core i7-4790 processors. The detailed comparison of on-chip resources of the FPGA platforms is presented in Fig.~\ref{fig:resource}. We use Xilinx SDX 2017.4 as the commercial high-level synthesis backend to synthesize the high-level (C/C++) based RNN designs on the selected FPGAs. The REQ-YOLO framework of FPGA implementation is operating at 200MHz.
For the tiny YOLO network, the performance of the first four layers is bound by the communication due to the large input/output feature map size. The last five layers of the YOLO network are otherwise constrained by the computation because of the increased channel size. 

We conduct the comparison between the heterogeneous-based YOLO quantization method and the equal-distance-based quantization method on the selected ADM-7v3 platform. We report the resource usage (percentage) of two methods in Fig.~\ref{fig:resource}. We can observe that the heterogeneous-based method better exploit the hardware resource than the equal-distance method, especially in LUT, therefore leading to higher performance and throughput. This finding also verifies our discussion in Section~\ref{sec:resouce-aware}.  
The layer-wise computation, communication and latency analysis of both equal-distance quantization and heterogeneous quantization is shown in Table~\ref{tbl:yolo}. In the communication-bound layers, the latency is the same for both methods. Overall, the heterogeneous-based method achieves 1.5$\times$ performance compared to the equal-distance method.

The results of performance and energy efficiency of our FPGA based YOLO implementations are presented in Table~\ref{tb1:platform}. Our FPGA-YOLO1 using heterogeneous quantization outperforms our FPGA-YOLO0 using equal-distance quantization in terms of both performance and energy efficiency, i.e., 1.5$\times$ in performance and 1.6$\times$ in energy efficiency, since the heterogeneous quantization fully exploits the hardware resource and design parallelism. Please note that since the DataDJI dataset is the latest released, we cannot find the related FPGA based implementations to compare with. For PASCAL VOC dataset, compared to other FPGA based works~\cite{guo2016model,ma2017hardware}, our FPGA-YOLO1 achieves at least 10$\times$ performance enhancement, while the FPGA fabric Virtex-7 690t in our platform is only slightly better than Virtex-7 485t used in~\cite{ma2017hardware} in resource capacity. We can not compare the energy efficiency among them since the power measurements are not provided in~\cite{guo2016model,ma2017hardware}.

\subsubsection{Cross-platform Comparison}
We implement the same YOLO network on two GPU platforms and compare with the tiny YOLO proposed in~\cite{redmon2016you} using Titan X GPU.
The first one is GeForce GTX 1070, which is a Nvidia GPU designed for PC. The second one is a Jetson TX2, which is the latest embedded GPU platform. The detailed specifications and comparisons among these platforms are shown in Table~\ref{tb1:platform}. We implement the trained model on both platforms and measure the performance using frame per second (FPS) and power consumption (W). Compared to Titan X-YOLO~\cite{redmon2016you}, our GTX-YOLO and our TX2-YOLO achieve 1.8$\times$ and 2.9$\times$ enhancement in energy efficiency.

Compared to GPU-based YOLO implementation (Our GTX-YOLO), our two FPGA YOLO implementations has the similar or better speed while dissipating around 6$\times$ less power, and the efficiency (performance per power) of our FPGA-YOLO0 and our FPGA-YOLO1 are 5.7$\times$ and 9.4$\times$ better, respectively. It indicates that our proposed REQ-YOLO framework is very suitable for FPGAs, since usually GPUs often perform faster than FPGAs as discussed in~\cite{betkaoui2010comparing}. Compared to the GPU-based YOLO implementation with the best energy efficiency (our TX2-YOLO), our two FPGA YOLO implementations achieve 3.5$\times$ and 5.8$\times$ improvement in energy efficiency. While our FPGA YOLO implementations are at least 7.3$\times$ faster while only dissipating at most 2.1$\times$ more power. 

Overall, our proposed REQ-YOLO framework is effective on both GPUs and FPGAs. It is highly promising to deploy our proposed REQ-YOLO framework on FPGA to gain much higher energy efficiency for autonomous systems on object sections than on GPUs. More importantly, the proposed framework achieves much higher FPS over the real-time requirement.


\section{Conclusion}
In this work, we propose REQ-YOLO, a resource-aware, systematic weight quantization framework for object detection, considering both algorithm and hardware resource aspects in object detection. We adopt the block-circulant matrix method and we incorporate ADMM with FFT/IFFT and develop a heterogeneous weight quantization method including both equal-distance and heterogeneous quantization methods considering hardware resource.
We implement the quantized models on the state-of-the-art FPGA taking advantage of the potential to store the whole compressed DNN models on-chip. To achieve real-time, highly-efficient implementations on FPGA, we develop an efficient PE structure supporting both equal-distance and mixed powers-of-two quantization methods, CONV dataflow and pipelining techniques, design optimization techniques focus on reducing memory access and PE size/numbers, and a template-based automatic synthesis framework to optimally exploit hardware resource. Experimental results show that our proposed framework can significantly compress the YOLO model while introducing very small accuracy degradation. Our framework is very suitable for FPGA and our FPGA implementations outperform the state-of-the-art designs.

\begin{acks}
This work is supported by Beijing Natural Science Foundation (No. L172004), Municipal Science and Technology Program under Grant Z181100008918015, and National Science Foundation under grants CNS \#1704662 and CNS \#1739748. We thank all the anonymous reviewers for their feedback.
\end{acks}

\bibliographystyle{ACM-Reference-Format}
\bibliography{FPGA2019REQYOLO}

\end{document}